\documentclass{article}

\usepackage[dblblindworkshop, final]{neurips_2026}

\usepackage[utf8]{inputenc}
\usepackage[T1]{fontenc}
\usepackage{hyperref}
\usepackage{url}
\usepackage{booktabs}
\usepackage{amsfonts}
\usepackage{amsmath}
\usepackage{nicefrac}
\usepackage{microtype}
\usepackage{xcolor}
\usepackage{graphicx}
\usepackage{subcaption}
\usepackage{multirow}
\usepackage{tabularx}

\title{LM6-CastBench: A Geometry-Aware Simulation Dataset for Transient Thermal Prediction in LM6 Sand Casting}
\workshoptitle{AIDaR: AI Data Readiness for Scientific Discovery}

\author{%
  Suhani Kolhatkar \\
  COEP Technological University \\
  \And
  Anand Khandekar \\
  Arcticturn Foundation \\
}

\begin{document}

\maketitle

\begin{abstract}
Transient thermal fields during solidification connect casting process conditions and cooling behaviour to the formation of defect-prone regions in aluminum castings. Repeatedly solving this thermal problem across geometries and process conditions is computationally expensive, motivating data-driven surrogate models. However, existing machine-learning studies on casting thermal behaviour rely on study-specific simulation databases that do not support reproducible geometry-level benchmarking. We introduce \textbf{LM6-CastBench}, an open simulation dataset for geometry-aware learning of transient thermal fields in LM6 aluminum sand casting, combining three geometry families, ten variants per family, five pouring temperatures, and three mould temperatures (450 geometry-process cases in total). Each case retains CAD geometry, unstructured mesh connectivity, spatial descriptors, process metadata, and time-resolved nodal temperatures. We benchmark a point-wise MLP against a spatially structured U-Net under geometry-level data splits; the U-Net reduces temperature-field errors by a large margin, with the biggest gains on geometries with localized features and abrupt section changes. LM6-CastBench gives researchers a reproducible dataset and benchmark for geometry-aware surrogate modelling of casting thermal behaviour.
\end{abstract}

\section{Introduction}
\label{sec:intro}

Transient thermal behaviour governs solidification and is an important factor in the formation of defects such as porosity and shrinkage-related discontinuities \cite{samuel2023porosity,campbell2015casting}; the dataset and models in this paper target the transient temperature field under a conduction-dominated assumption and do not simulate mould filling or predict defects directly. Process conditions such as pouring temperature further affect solidification behaviour in LM6 aluminium alloy \cite{ahmad2013pouring}. Finite-element simulation resolves these fields \cite{verma2018lm6,chen2022casting}, but repeated simulation across geometries and process conditions is computationally demanding, motivating surrogate models for rapid prediction \cite{bhosekar2018surrogate}.

Recent work has learned casting thermal behaviour directly from simulation data: Kang et al.~\cite{kang2024deeplearning,kang2025diffusion} use procedurally generated 2D geometries with image-like temperature fields, and Ktari et al.~\cite{ktari2026neuralnetwork} retain finite-element nodal temperatures for a fixed casting configuration while varying only pouring temperature. Neither provides a benchmark where 3D geometry and process variation are both systematically controlled while the underlying mesh and transient field remain available; reduced-order and mesh-based casting approaches face the same limitation \cite{zhang2026digitaltwin}.

LM6-CastBench addresses this gap. It provides three geometry families (brake lever, flange, stepped block) with ten controlled variants each, simulated over five pouring and three mould temperatures (450 geometry-process cases), while keeping geometry, mesh, process conditions, and transient thermal response linked in a common representation rather than releasing flattened temperature tensors. To demonstrate the value of this organization, we compare a point-wise MLP against a spatially structured 3D U-Net under geometry-level splits, testing whether the retained spatial structure has a measurable effect on prediction and providing a baseline for future geometry-aware surrogates.

Our contributions are threefold: (1) a controlled 3D simulation dataset with 450 transient thermal cases spanning geometry and process variation; (2) a geometry- and process-aware data organization enabling reproducible geometry-level evaluation; (3) a benchmark showing that spatially structured prediction substantially improves thermal-field reconstruction, especially for geometries with localized features and abrupt section changes.

\section{Dataset Generation}
\label{sec:pipeline}

\begin{figure}[t]
  \centering
  \includegraphics[width=0.65\linewidth]{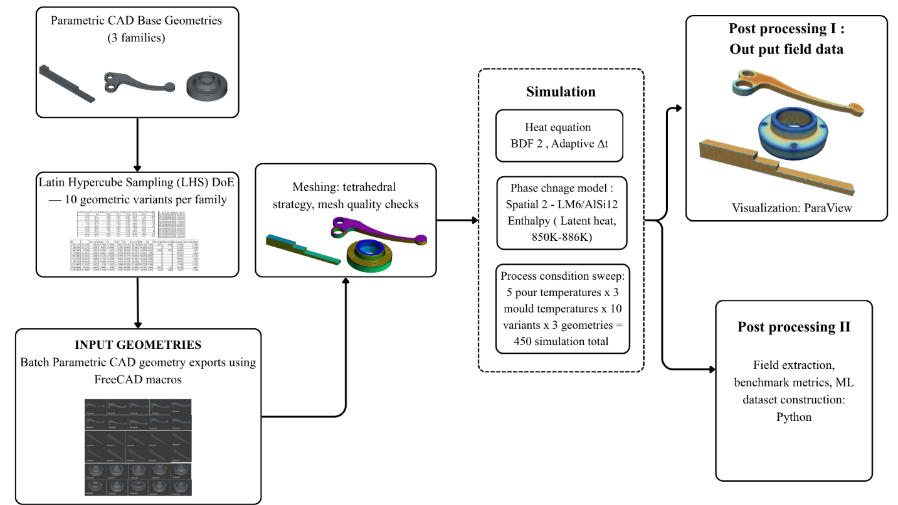}
  \caption{End-to-end workflow for generating LM6-CastBench.}
  \label{fig:workflow}
\end{figure}

LM6-CastBench is built as an end-to-end simulation-to-dataset pipeline (Figure~\ref{fig:workflow}): controlled geometric variation, a shared process-condition grid across every geometry, the transient thermal solve, and preservation of the resulting fields together with their spatial and process descriptors.

Three geometry families (brake lever, flange, stepped block) are each parametrized with ten LHS-sampled variants \cite{mckay1979lhs}, exported as STEP models, and independently meshed with first-order tetrahedral elements (quality-checked; local refinement near holes, fillets, and curved features). Each variant is then simulated under all $5 \times 3$ pouring/mould temperature combinations using an enthalpy-based finite-element formulation in Elmer FEM \cite{malinen2013elmer}, giving 450 geometry-process cases in total. Full parameter ranges, meshing settings, and solver configuration are given in Appendices~\ref{app:A} and \ref{app:B}.

For each case, the released dataset retains node coordinates, mesh connectivity, signed-distance values, process parameters, and time-resolved nodal temperatures, so that geometry and its corresponding thermal response are never separated. Data are split 80/10/10 (train/val/test) at the \emph{geometry-variant} level within each family -- 8 of the 10 variants per family for training, 1 for validation, and 1 for test (3 held-out test geometries in total) -- so timesteps from the same physical geometry never appear in both training and test sets; we note that three held-out geometries is a small test set and treat the benchmark (Section~\ref{sec:benchmark}) as illustrative of the dataset's utility rather than a statistically exhaustive evaluation.

\paragraph{Data readiness design.} Each case carries a unique (geometry-family, variant, process-condition) identifier and a metadata record linking this identifier to its CAD, mesh, and field file paths, its process parameters, and its split assignment -- a provenance chain (field $\rightarrow$ mesh $\rightarrow$ CAD $\rightarrow$ process $\rightarrow$ solver setting) that lets a user or agent query, filter, or regroup cases, and stratify errors by any of these fields, without re-deriving them from raw solver output. The fixed split manifest, split seed, and benchmark scripts are released with the repository so future methods can be compared against the same partitions.

\begin{figure}[t]
  \centering
  \includegraphics[width=\linewidth]{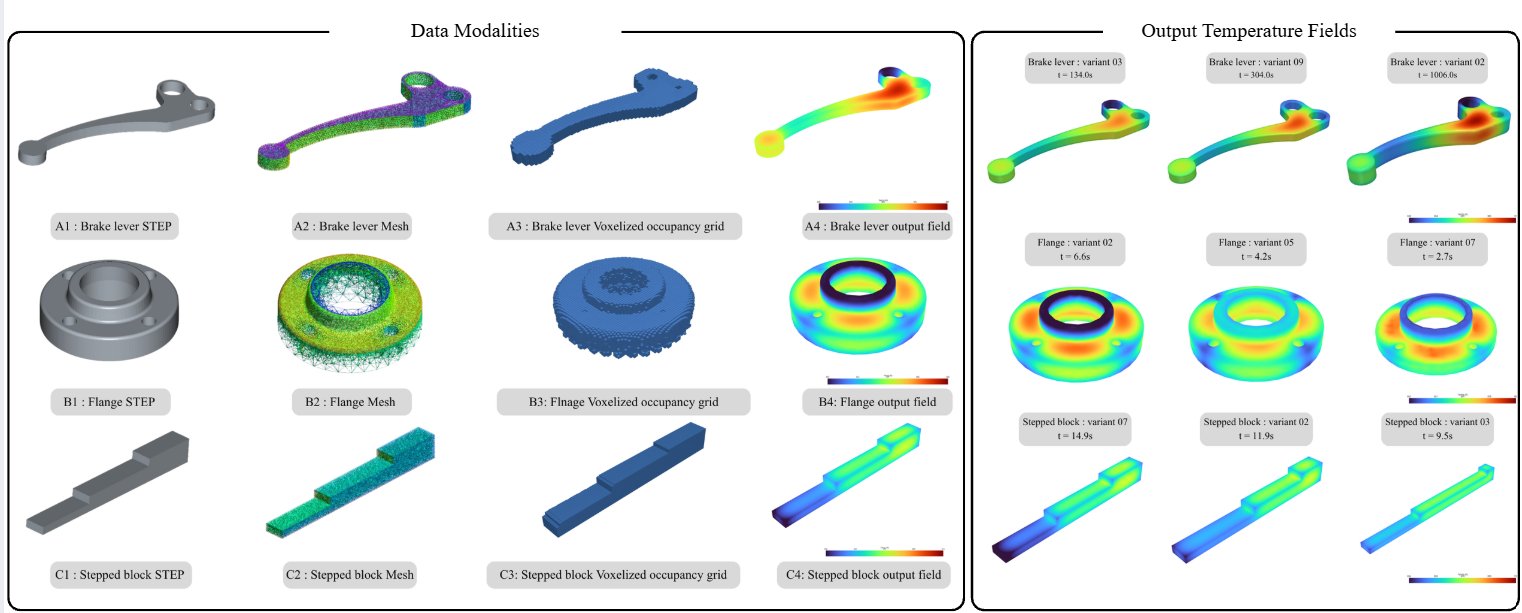}
  \caption{Geometry, spatial representations, and transient thermal fields in LM6-CastBench.}
  \label{fig:geometry}
\end{figure}

Figure~\ref{fig:geometry} shows representative variants across the three families together with their CAD geometry, mesh, voxelized occupancy representation, and simulated temperature snapshots. The repository contains all 450 cases as preprocessed PyTorch tensor (\texttt{.pt}) files, the generation and benchmark scripts, released under \texttt{CC-BY-4.0}: \url{https://github.com/SuhaniKolhatkar/LM6-CastBench}.

\section{ML Benchmarking}
\label{sec:benchmark}

We compare two surrogate formulations with different spatial representations under the same geometry-level splits: a point-wise MLP predicting nodal temperature from coordinates, signed-distance value, process temperatures, and time; and a grid-based 3D U-Net \cite{ronneberger2015unet} predicting the field on a $32\times32\times32$ voxel grid from occupancy, signed-distance, process, and temporal channels, mapped back to FEM nodes for evaluation (Table~\ref{tab:implementation}). Both are trained with MSE/Adam and evaluated with NMSE, MAE, relative $L_2$, P95 error, temporal correlation, and inference time. For each test case, node, and time, we compute the absolute error between predicted and reference temperature; MAE, NMSE, and relative $L_2$ are averaged over all test nodes, timesteps, and geometry-process cases, P95 is the 95th percentile of the resulting absolute-error distribution, and temporal correlation is the Pearson correlation between predicted and reference nodal time series, averaged over nodes and cases. Implementation and training details (parameter counts, epochs, batch size, learning rate, hardware) are given in Appendix~\ref{app:B}.

\begin{table}[t]
  \caption{Benchmark implementation details of the point-wise MLP and grid-based 3D U-Net.}
  \label{tab:implementation}
  \centering
  \small
  \begin{tabular}{lll}
    \toprule
    \textbf{Implementation} & \textbf{Point-wise MLP} & \textbf{Grid-based 3D U-Net} \\
    \midrule
    Representation & FEM nodes & $32^3$ voxel grid \\
    Input & $X,Y,Z,SDF,T_{pour},T_{mould},t$ & Occupancy, SDF, $T_{pour},T_{mould},t$ \\
    Output & Nodal temperature & 3D temperature field \\
    Architecture & 3-hidden-layer MLP & 3D encoder--decoder U-Net \\
    \bottomrule
  \end{tabular}
\end{table}

\textbf{Quantitative results.} Table~\ref{tab:results} summarizes performance on held-out geometry variants. The U-Net consistently reduces error relative to the MLP, modestly for the brake geometry and substantially for flange and stepped block: MAE falls from 8.38~K to 2.19~K (brake), 36.75~K to 1.03~K (flange), and 80.71~K to 1.86~K (stepped block), with P95 error following the same pattern. The MLP is faster at inference (0.51-1.20~s vs.\ 2.66-8.04~s), so the benchmark separates computational cost from prediction fidelity.

\begin{table}[t]
  \caption{Quantitative temperature-field prediction performance of the MLP and U-Net on held-out geometry variants.}
  \label{tab:results}
  \centering
  \small
  \begin{tabular}{llrrrrrr}
    \toprule
    \textbf{Geometry} & \textbf{Model} & \textbf{NMSE} & \textbf{MAE (K)} & \textbf{Rel.\ $L_2$} & \textbf{P95 (K)} & \textbf{Temp.\ Corr.} & \textbf{Time (s)} \\
    \midrule
    Brake         & MLP    & 0.002456 & 8.382  & 0.0149 & 21.125  & 0.9996 & 0.511 \\
    Brake         & U-Net  & 0.000200 & 2.187  & 0.0045 & 6.817   & 1.0000 & 2.662 \\
    Flange        & MLP    & 0.0642   & 36.748 & 0.0626 & 66.151  & 0.9961 & 1.195 \\
    Flange        & U-Net  & 7.61E-05 & 1.033  & 0.0023 & 2.431   & 1.0000 & 4.516 \\
    Stepped Block & MLP    & 0.2543   & 80.707 & 0.1328 & 112.097 & 0.9986 & 1.065 \\
    Stepped Block & U-Net  & 3.08E-04 & 1.861  & 0.0055 & 5.061   & 0.9999 & 8.041 \\
    \bottomrule
  \end{tabular}
\end{table}

\begin{figure}[t]
  \centering
  \begin{subfigure}[t]{0.49\linewidth}
    \centering
    \includegraphics[width=\linewidth]{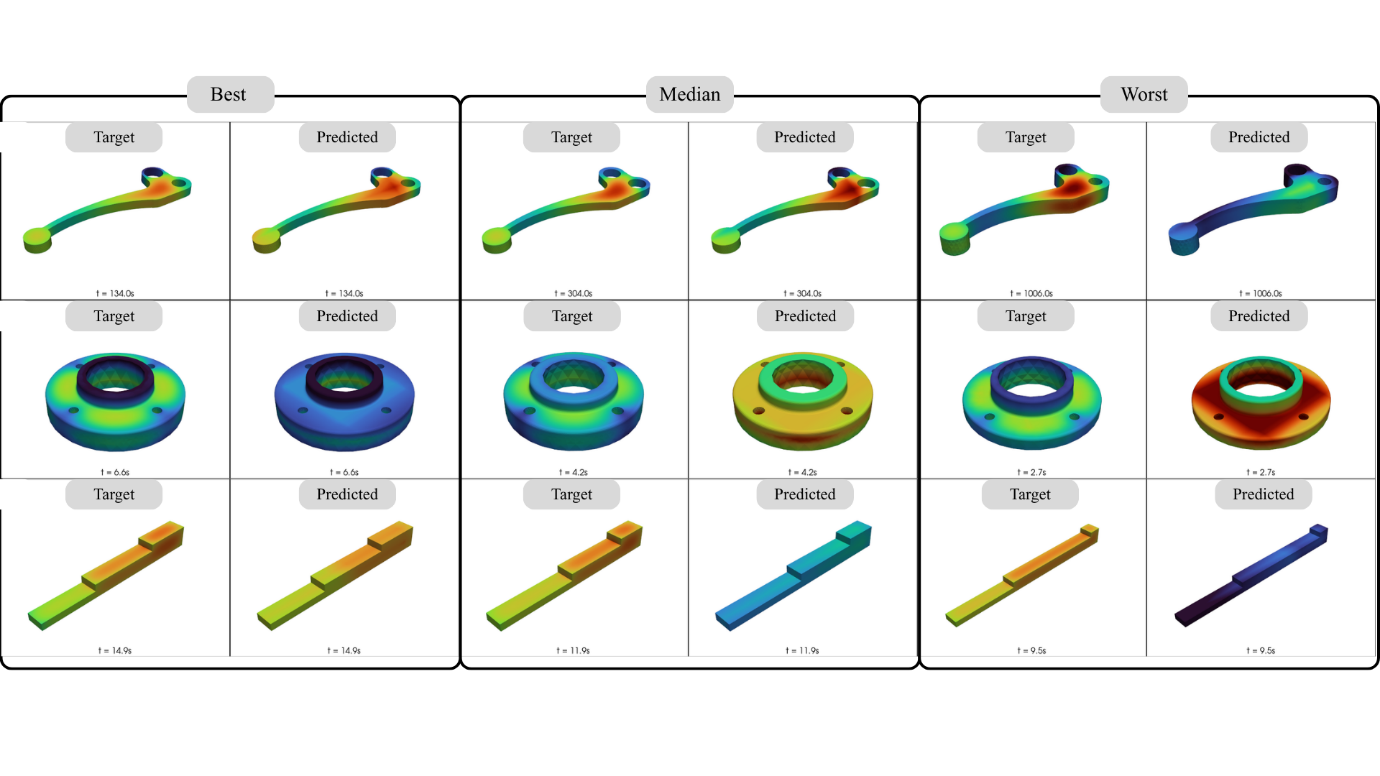}
    \caption{Point-wise MLP: predictions and error maps across representative LM6-CastBench geometries.}
    \label{fig:mlp_pred}
  \end{subfigure}
  \hfill
  \begin{subfigure}[t]{0.49\linewidth}
    \centering
    \includegraphics[width=\linewidth]{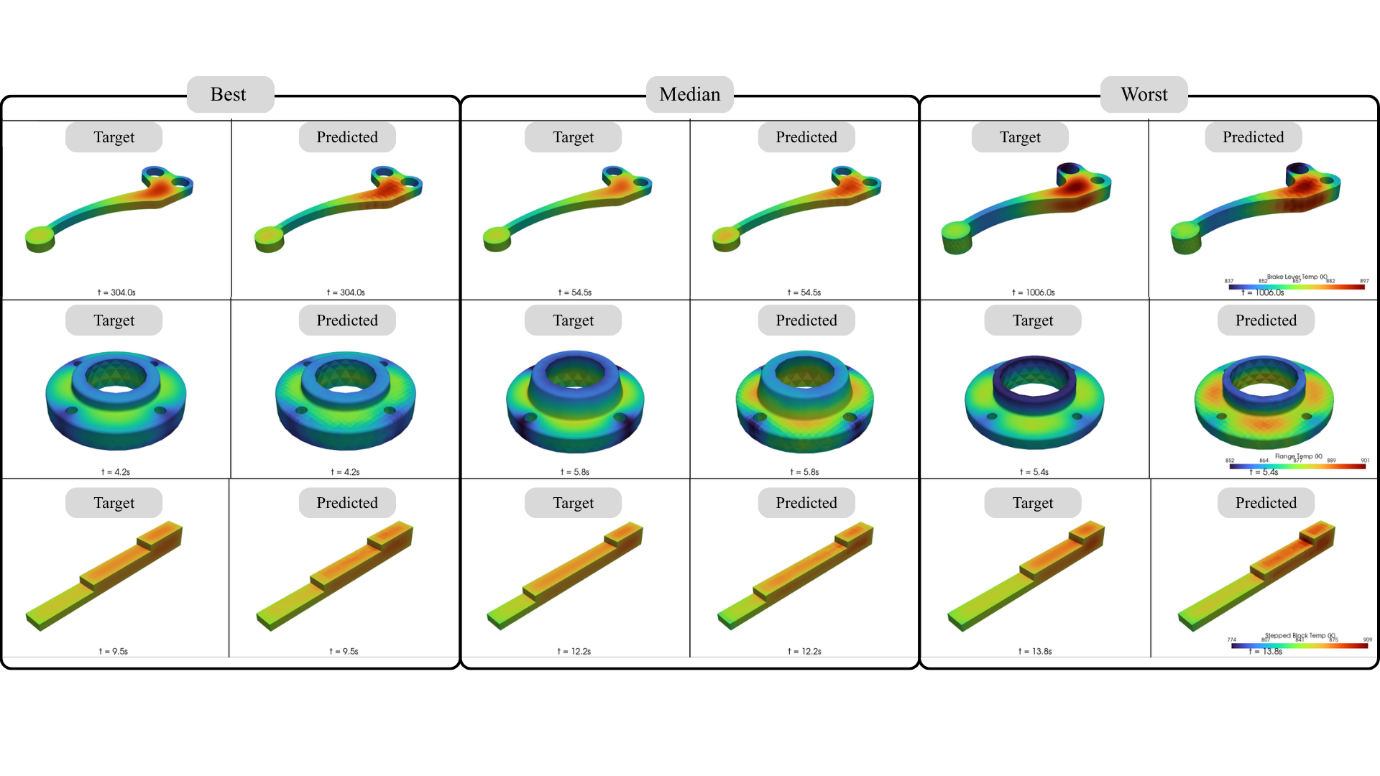}
    \caption{Grid-based 3D U-Net: predictions and error maps across representative LM6-CastBench geometries.}
    \label{fig:unet_pred}
  \end{subfigure}
  \caption{Transient temperature-field predictions and error maps for the point-wise MLP (left) and grid-based 3D U-Net (right) across representative LM6-CastBench geometries.}
  \label{fig:mlp_unet_pred}
\end{figure}

\textbf{Spatial behaviour.} Figure~\ref{fig:mlp_unet_pred} compares predicted fields and absolute-error maps across the three geometries. The MLP shows larger, more persistent errors around the flange ring, bolt-hole regions, and the stepped block's cross-sectional transitions, consistent with its point-wise (no neighbouring-context) representation; the U-Net tracks the simulated field more closely in these regions by processing neighbouring spatial locations jointly. Table~\ref{tab:results} reports errors averaged over all test geometries, process conditions, nodes, and timesteps, whereas Appendix~\ref{app:C} reports individual best/median/worst cases, which show larger instantaneous errors during the early transient but are consistent with these aggregate trends. This indicates that the geometric and spatial structure preserved in LM6-CastBench is consequential for prediction, and gives a consistent basis for evaluating further surrogates (e.g., graph networks, neural operators) against the same held-out geometries.

\section*{Limitations and Future Scope}

The dataset currently covers ten variants per family and two process parameters (pouring and mould temperature); broader geometric and process coverage, parametrized gating/riser systems, and coupled filling-solidification physics are natural extensions. The U-Net's accuracy is also bounded by the $32^3$ voxel quantisation, since mesh nodes sharing a voxel receive the same prediction. The benchmark currently uses only two baseline architectures, and the dataset is intended to support broader comparisons with graph networks, neural operators, and physics-informed models \cite{pfaff2021meshgraph,li2021fno,kushwaha2024deeponet,raissi2019pinn,lu2021deepxde}.

\section{Conclusion}
\label{sec:conclusion}

LM6-CastBench links geometry, process conditions, mesh discretization, and transient thermal response in a single reproducible dataset of 450 cases across three geometry families. The MLP/U-Net benchmark shows that this retained spatial structure has a measurable, geometry-dependent effect on prediction accuracy, largest where geometries contain localized features and abrupt section changes. Beyond this comparison, the dataset offers a controlled space for evaluating future geometry-aware surrogate and neural-operator methods on the same underlying physics.

\paragraph{Large Language Model Disclosure.} Large language models were used only to review, check, and verify author-written model implementation codes; all codes, modelling decisions, and final results were developed and validated by the authors.

\newpage

\bibliographystyle{plain}
\bibliography{references}

\newpage

\appendix

\section{CAD Parametrization and Geometry Generation}
\label{app:A}

The first component of the dataset is controlled geometric diversity. LM6-CastBench contains three casting geometry families -- brake lever, flange, and stepped block -- each represented by ten geometrically distinct variants generated from a fixed set of parametrization rules while preserving the topology of the family. Rather than generating arbitrary shapes, independent geometric parameters are sampled within their design spaces using Latin Hypercube Sampling (LHS) \cite{mckay1979lhs}, covering the prescribed parameter space without a full factorial geometry study and retaining a recognizable relationship between variants within each family. The resulting CAD models are exported in STEP format and passed independently through the downstream meshing and simulation stages; each variant remains an identifiable geometry-level unit throughout the dataset, which makes it possible to separate geometries during evaluation and prevents different time samples from the same physical shape from being treated as independent geometries.

\begin{figure}[t]
  \centering
  \includegraphics[width=\linewidth]{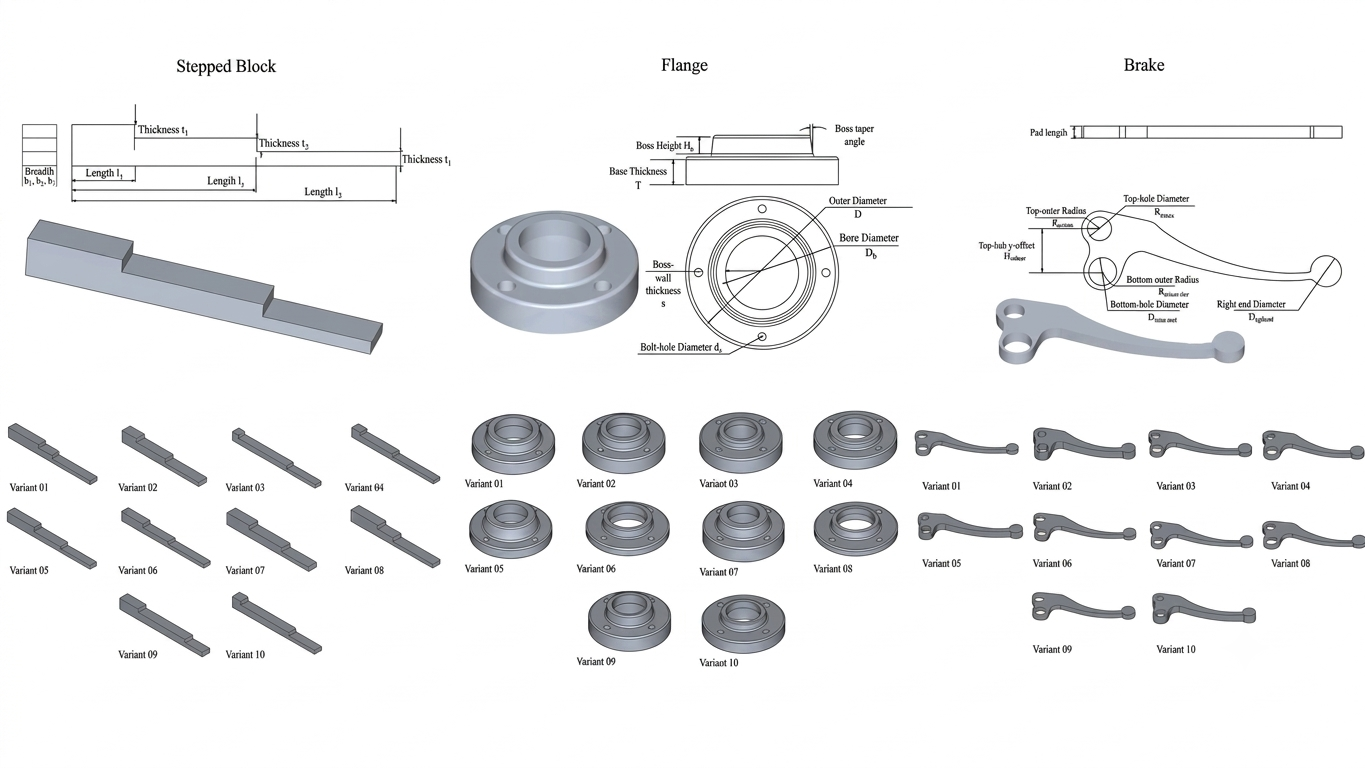}
  \caption{CAD Parametrization details}
  \label{fig:parametrization}
\end{figure}

Table~\ref{tab:paramsA1} summarizes the design parameters for each family, their definitions or governing relations, and the corresponding ranges or fixed values used for CAD variant generation.

\begin{table}[t]
  \caption{Parametric design variables used to generate the three LM6 casting geometry families. The table lists the independent geometric parameters, their definitions or governing relations, and the corresponding design ranges or fixed values used during Latin Hypercube Sampling.}
  \label{tab:paramsA1}
  \centering
  \small
  \begin{tabularx}{\textwidth}{@{}l l X l@{}}
    \toprule
    \textbf{Family} & \textbf{Parameter} & \textbf{Definition / formula} & \textbf{Range or value} \\
    \midrule
    \multirow{5}{*}{Stepped block}
      & $l_1$ & Length, layer 1 (DOE) & 470-670 mm \\
      & $l_2$ & Length, layer 2 (DOE) & 250-450 mm \\
      & $l_3$ & Length, layer 3 (DOE) & 30-230 mm \\
      & $b_1,b_2,b_3$ & Breadth, layers 1-3 (fixed) & 50 mm \\
      & $t_1,t_2,t_3$ & Thickness, layers 1-3 (fixed) & 20 mm \\
    \midrule
    \multirow{9}{*}{Flange}
      & $D$ & Outer diameter (DOE) & 80-140 mm \\
      & $T$ & Base thickness (DOE) & 10-25 mm \\
      & $D_b/D$ & Bore ratio (DOE) & 0.42-0.52 \\
      & $t_{wall}$ & Boss-wall thickness (DOE) & 6-10 mm \\
      & $H_b$ & Boss height (DOE) & 8-20 mm \\
      & --- & Boss taper angle (DOE) & $-15^\circ$ to $0^\circ$ \\
      & $d_h$ & Bolt-hole diameter (DOE) & 6-10 mm \\
      & $D_b$ & Bore diameter (derived): $D_b = (D_b/D)\cdot D$ & derived \\
      & $D_{boss}$ & Boss outer diameter (derived): $D_{boss} = D_b + 2\cdot t_{wall}$ & derived \\
      & --- & 4 bolt holes on the bolt circle between boss and rim, at orthogonal positions & --- \\
    \midrule
    \multirow{8}{*}{Brake lever}
      & $D_{bottom,hole}$ & Bottom-hole diameter (DOE, anchor) & 12-16 mm \\
      & $D_{top,hole}$ & Top-hole diameter (DOE, anchor) & 10-14 mm \\
      & $R_{bottom,outer}$ & Bottom outer radius (DOE, anchor) & 9.5-11.5 mm \\
      & $R_{top,outer}$ & Top outer radius (DOE, anchor) & 9-11 mm \\
      & $D_{right,end}$ & Right-end diameter (DOE, anchor) & 14-18 mm \\
      & $L_{span,X}$ & Right-end $x$-offset from bottom anchor (DOE) & -- \\
      & $H_{offset,Y}$ & Top-hub $y$-offset (DOE, anchor) & 18-28 mm \\
      & --- & Pad length (DOE) & 6-12 mm \\
      & --- & Bottom anchor fixed as reference; top and right-end anchors repositioned by $H_{offset,Y}$ and $L_{span,X}$; all other sketch points (line ends, fillet centres, B\'ezier poles) displaced by inverse-distance-weighted (IDW) interpolation of the three anchor displacements, so originally coincident points remain coincident & --- \\
    \bottomrule
  \end{tabularx}
\end{table}

\section{Meshing Strategy, Simulation Methodology and Material Properties}
\label{app:B}

\subsection{Meshing Strategy}

Each geometry variant is independently meshed using first-order tetrahedral elements, with local refinement applied around geometrically complex regions such as holes, fillets, and curved features. Rather than deforming a common base mesh to represent different variants, each mesh is generated independently from its CAD geometry, so it retains the spatial differences introduced during parametrization; mesh-quality checks are applied across all three geometry families before thermal simulation so that released temperature fields remain associated with their original unstructured spatial discretization rather than being reduced to temperature values alone.

Each STEP geometry was meshed independently, with a dedicated mesh generated for every geometry instance rather than by morphing or deforming a common base mesh, preserving the geometric fidelity of individual variants and allowing local mesh refinement to follow geometry-specific features. First-order tetrahedral elements were used throughout the dataset. Surface meshing was performed using the Frontal-Delaunay algorithm, followed by volumetric meshing using the HXT algorithm.

\paragraph{Mesh Quality.} Mesh quality was evaluated for all geometry variants using standard quality measures. The target ranges maintained during mesh generation are summarized in Table~\ref{tab:meshquality}.

\begin{table}[h]
  \caption{Target mesh-quality ranges used for LM6-CastBench.}
  \label{tab:meshquality}
  \centering
  \begin{tabular}{ll}
    \toprule
    Mesh-quality metric & Target range \\
    \midrule
    Signed Inverse Condition Number (SICN) & 0.87-0.90 \\
    Gamma & 0.80-0.90 \\
    Signed Inverse Gradient Error (SIGE) & 0.10-0.40 \\
    \bottomrule
  \end{tabular}
\end{table}

\paragraph{Element Sizing Strategy.} Element size was selected according to the characteristic dimensions of each geometry rather than using a single global mesh size. The principal geometric measures considered were the overall bounding-box dimensions, hole and fillet diameters, and minimum wall thickness. This geometry-dependent sizing strategy was complemented by curvature-based refinement in regions with significant geometric variation, including bores, fillets, and bolt holes. Consequently, finer elements were generated in curved and geometrically complex regions, while relatively coarser elements were retained in planar and geometrically simple regions.

\paragraph{Physical Groups.} Each geometry was defined using two physical groups: one physical volume representing the complete casting domain and one physical surface representing the complete external boundary. The volume group was used for material-property assignment, while the surface group provided the boundary region required for application of the thermal boundary conditions during simulation.

\begin{table}[h]
  \caption{Physical groups defined for each meshed geometry.}
  \label{tab:physicalgroups}
  \centering
  \begin{tabular}{lll}
    \toprule
    Physical group & Region represented & Purpose \\
    \midrule
    Casting volume & Complete casting domain & Material-property assignment \\
    External surface & Complete external boundary & Application of thermal boundary conditions \\
    \bottomrule
  \end{tabular}
\end{table}

\subsection{Process-Condition Design}

Every geometry variant is exposed to the same process-condition grid: five pouring temperatures and three mould temperatures, giving 15 thermal simulation cases per geometry variant and $3 \times 10 \times 15 = 450$ cases in total (Table~\ref{tab:process}). This design separates process variation from geometry variation and ensures balanced representation of process conditions throughout the dataset.

\begin{table}[htbp]
  \caption{Process-condition design of LM6-CastBench.}
  \label{tab:process}
  \centering
  \begin{tabular}{ll}
    \toprule
    \textbf{Parameter} & \textbf{Values} \\
    \midrule
    Pouring temperature, $T_{pour}$ & 953 K, 973 K, 993 K, 1013 K, 1033 K \\
    Mould temperature, $T_{mould}$ & 303 K, 353 K, 403 K \\
    Process combinations per geometry & $5 \times 3 = 15$ \\
    Geometry variants & 3 families $\times$ 10 variants \\
    Total simulation cases & $3 \times 10 \times 15 = 450$ \\
    \bottomrule
  \end{tabular}
\end{table}

\subsection{Governing Thermal Model}

The transient thermal simulations were performed using the Elmer FEM solver. Heat transfer within the casting was modelled under a conduction-dominated assumption, with solidification represented using an enthalpy-based phase-change formulation; metal flow and convective transport within the molten alloy were not considered in the present baseline model. Simulations were performed in Cartesian coordinates with a second-order backward differentiation (BDF2) scheme for temporal discretization, an enthalpy-based phase-change model, a linear-system tolerance of $1\times10^{-6}$ for the heat equation and $1\times10^{-7}$ for the flux/gradient solve, and a nonlinear-system tolerance of $1\times10^{-5}$. The initial condition was a spatially uniform temperature field at the pouring temperature ($T = T_{pour}$). Time stepping used a geometry-specific schedule of progressively coarsening steps (fine steps during the early transient, coarser steps as cooling slows): $600\times0.1$~s + $400\times0.5$~s + $400\times2.0$~s + $400\times20.0$~s for the brake and stepped-block geometries, and $600\times0.1$~s + $240\times1.0$~s + $100\times5.0$~s + $80\times20.0$~s for the flange geometry. For each geometry variant, the process space was sampled by combining five pouring-temperature levels with three mould-temperature levels, resulting in 15 thermal simulation cases per geometry variant.

The transient thermal field was governed by the enthalpy-based energy equation
\begin{equation}
  \rho\, c_{eff}(T) \frac{\partial T}{\partial t} = \nabla \cdot \left[ k(T) \nabla T \right],
  \label{eq:energy}
\end{equation}
where $\rho$ is the material density, $c_{eff}(T)$ is the effective heat capacity incorporating the phase-change contribution, and $k(T)$ is the temperature-dependent thermal conductivity.

Heat transfer between the casting and mould was represented using a temperature-dependent convective (Robin-type) boundary condition,
\begin{equation}
  -k(T) \nabla T \cdot n = h(T)\left[T - T_{ext}\right],
  \label{eq:bc}
\end{equation}
where $h(T)$ denotes the temperature-dependent heat-transfer coefficient (Table~\ref{tab:ht}) and $T_{ext} = 353$~K is the prescribed external/ambient temperature used in the boundary condition.

\subsection{Material Properties}

The casting material was modelled as LM6 aluminium alloy (Al--Si12) with a constant density $\rho = 2650$~kg/m$^3$ (Table~\ref{tab:material}). Tables~\ref{tab:cp}-\ref{tab:ht} list the temperature-dependent specific heat $c_p(T)$, thermal conductivity $k(T)$, and convective boundary coefficient $h(T)$ exactly as specified in the Elmer \texttt{.sif} solver input file (piecewise-linear between listed points); the full \texttt{.sif} files are provided in the repository. The enthalpy table used for the phase-change (latent-heat) treatment has its onset near $850$~K and plateaus near $886$~K; treating these as the effective solidus and liquidus temperatures gives $T_{sol}\approx850$~K and $T_{liq}\approx886$~K, with a latent heat of fusion $L\approx3.9\times10^{5}$~J/kg obtained from the enthalpy change across this interval. These three values are \emph{inferred/derived} from the enthalpy curve rather than set as explicit standalone keywords in the \texttt{.sif} file, and are reported here for transparency; the source enthalpy table is included in the repository so they can be independently recomputed.

\begin{table}[h]
  \caption{General material and boundary-condition properties of LM6 used in the simulations.}
  \label{tab:material}
  \centering
  \small
  \begin{tabular}{ll}
    \toprule
    \textbf{Property} & \textbf{Value} \\
    \midrule
    Density, $\rho$ & 2650 kg/m$^3$ (constant) \\
    Solidus temperature, $T_{sol}$ (inferred) & $\approx 850$ K \\
    Liquidus temperature, $T_{liq}$ (inferred) & $\approx 886$ K \\
    Latent heat of fusion, $L$ (derived) & $\approx 3.9\times10^{5}$ J/kg \\
    Initial condition & Uniform, $T = T_{pour}$ \\
    External/ambient temperature, $T_{ext}$ & 353 K \\
    Phase-change treatment & Enthalpy-based \\
    \bottomrule
  \end{tabular}
\end{table}

The specific heat capacity is defined as:

\begin{table}[h]
  \caption{Specific heat $c_p(T)$ of LM6, from the Elmer \texttt{.sif} solver input file (piecewise-linear between listed points).}
  \label{tab:cp}
  \centering
  \begin{tabular}{cc}
    \toprule
    \textbf{Temperature (K)} & \boldmath$c_p$ \textbf{(J\,kg$^{-1}$\,K$^{-1}$)} \\
    \midrule
    293  & 871  \\
    473  & 920  \\
    673  & 963  \\
    850  & 1050 \\
    886  & 1080 \\
    1000 & 1080 \\
    \bottomrule
  \end{tabular}
\end{table}

The thermal conductivity is defined as:

\begin{table}[h]
  \caption{Thermal conductivity $k(T)$ of LM6, from the Elmer \texttt{.sif} solver input file (piecewise-linear between listed points).}
  \label{tab:kt}
  \centering
  \begin{tabular}{cc}
    \toprule
    \textbf{Temperature (K)} & \boldmath$k$ \textbf{(W\,m$^{-1}$\,K$^{-1}$)} \\
    \midrule
    293  & 155 \\
    473  & 159 \\
    673  & 163 \\
    850  & 100 \\
    886  & 90  \\
    1000 & 90  \\
    \bottomrule
  \end{tabular}
\end{table}

The convective heat-transfer coefficient (Eq.~\eqref{eq:bc}) is defined as:

\begin{table}[h]
  \caption{Convective boundary coefficient $h(T)$, from the Elmer \texttt{.sif} solver input file (piecewise-linear between listed points).}
  \label{tab:ht}
  \centering
  \begin{tabular}{cc}
    \toprule
    \textbf{Temperature (K)} & \boldmath$h$ \textbf{(W\,m$^{-2}$\,K$^{-1}$)} \\
    \midrule
    293 & 35    \\
    850 & 110   \\
    868 & 300   \\
    886 & 633.3 \\
    993 & 633.3 \\
    \bottomrule
  \end{tabular}
\end{table}

\subsection{Numerical and Mesh Settings}

Table~\ref{tab:numerical} summarizes the solver and discretization settings; Table~\ref{tab:meshstats} reports mesh size for one representative variant per geometry family (node count, edge count, and number of saved timesteps, extracted with PyVista).

\begin{table}[htbp]
  \caption{Summary of numerical and process settings.}
  \label{tab:numerical}
  \centering
  \begin{tabular}{ll}
    \toprule
    \textbf{Parameter} & \textbf{Description} \\
    \midrule
    Analysis type & Transient thermal (Elmer FEM, Heat Equation) \\
    Time integration & BDF2 \\
    Linear tolerance (Heat Equation) & $1\times10^{-6}$ \\
    Linear tolerance (Flux/Gradient) & $1\times10^{-7}$ \\
    Nonlinear tolerance & $1\times10^{-5}$ \\
    Phase-change model & Spatial 2 (enthalpy-based) \\
    Pouring temperature & Five levels (953-1033 K) \\
    Mould temperature & Three levels (303-403 K) \\
    Cases per geometry variant & 15 \\
    Coordinate system & Cartesian \\
    Mesh-to-solver scaling & Applied consistently \\
    Linear solver & Iterative BiCGStab \\
    Flux/gradient solver & Iterative CG \\
    Elmer FEM version & ElmerSolver v26.1 \\
    Output & VTU temperature fields \\
    \bottomrule
  \end{tabular}
\end{table}

\begin{table}[htbp]
  \caption{Mesh statistics for one representative variant per geometry family.}
  \label{tab:meshstats}
  \centering
  \begin{tabular}{lrrr}
    \toprule
    \textbf{Geometry family} & \textbf{Nodes} & \textbf{Mesh edges} & \textbf{Saved timesteps} \\
    \midrule
    Brake         & 6{,}144  & 67{,}278  & 638 \\
    Flange        & 12{,}469 & 130{,}906 & 1{,}020 \\
    Stepped Block & 10{,}192 & 121{,}086 & 1{,}800 \\
    \bottomrule
  \end{tabular}
\end{table}

\subsection{Randomness, Seeds, and Licensing}

The geometry-variant train/validation/test split (Section~\ref{sec:pipeline}) uses a fixed random seed of \texttt{42} (\texttt{sklearn.train\_test\_split}), giving, concretely, 120/15/15 simulation cases per family for train/validation/test (8/1/1 geometry variants $\times$ 15 process conditions each). Model training used a single seed per model (not averaged over repeated seeds). The exact LHS geometry-sampling seed was not logged; since the generated STEP geometries for all 30 variants are released, the geometry set itself does not depend on it. The split-index manifest is included in the repository, and the dataset and code are released under the \texttt{CC-BY-4.0} licence. Environment and dependency versions are recorded in the repository to support exact reproduction.

\subsection{Model Training Details}

\textbf{MLP.} Input $(X,Y,Z,\mathrm{SDF},T_{pour},T_{mould},t)\in\mathbb{R}^7$; architecture \texttt{Linear(7$\to$128)-ReLU-Linear(128$\to$128)-ReLU-Linear(128$\to$64)-ReLU-Linear(64$\to$1)} with 25{,}857 trainable parameters; MSE loss on normalized values; Adam ($\beta_1=0.9$, $\beta_2=0.999$, $\epsilon=1\times10^{-8}$), learning rate $1\times10^{-3}$ (constant), batch size 4{,}096 (node, timestep) rows, 50 epochs; per-feature z-score input normalization and z-score target normalization (denormalized at inference), both using training-split statistics only.

\textbf{U-Net.} A separate U-Net was trained for each geometry family (brake, flange, stepped block). The three models share one architecture, one set of hyperparameters and one voxelisation; they differ only in their training data, and therefore in their normalisation statistics and best-validation epoch. Each variant is voxelised onto a $32\times32\times32$ grid over its node bounding box (5\% padding); each voxel takes the value of its nearest mesh node ($k$-d tree, \texttt{scipy.cKDTree}), and a voxel is marked occupied ($=1$) only if that node lies within 1.5 voxel diagonals, otherwise empty ($=0$). The 5 input channels are occupancy, SDF, $T_{pour}$, $T_{mould}$ and $t$ (the last three constant over the grid); the output is 1 channel (temperature). Grid-to-node inference uses a deterministic voxel-index lookup, so nodes sharing a voxel receive the same prediction. The network is a 3D U-Net with three down-sampling stages, a bottleneck and three up-sampling stages, with skip connections by concatenation at every level. Each convolution block is two repetitions of \texttt{Conv3d}($3^3$, padding 1)--\texttt{InstanceNorm3d}--ReLU. Encoder channels are $5\to16\to32\to64$ with \texttt{MaxPool3d} (factor 2) between stages, the bottleneck is $64\to128$, and the decoder uses \texttt{ConvTranspose3d} ($128\to64$, $64\to32$, $32\to16$), each followed by a convolution block after skip concatenation, ending in a $1\times1\times1$ \texttt{Conv3d} ($16\to1$). The model has 1{,}402{,}289 trainable parameters (about 1.4\,M), identical for all three geometries. Training uses MSE loss on normalized grid values, Adam with learning rate $1\times10^{-3}$ and no learning-rate schedule, batch size 8 (variant, timestep) grids, and 50 epochs. Inputs and target are z-score normalized per channel (occupancy excluded) using training-split statistics only. The best-validation checkpoint is used for evaluation. Because mesh nodes sharing a voxel receive identical predictions, the $32^3$ grid imposes a quantisation floor on U-Net accuracy even for a perfect model; the point-wise MLP, which operates at exact node positions, has no such floor.

\subsection{Hardware and Inference}

Training was performed on an NVIDIA Tesla T4 (16~GB VRAM, Kaggle T4$\times$2 instance); local inference timings (Table~\ref{tab:results}) were measured on an NVIDIA RTX 4050 Laptop GPU (6~GB VRAM). MLP inference is batched point-wise (batch size 8{,}192 rows) with the full field assembled on CPU; U-Net inference proceeds sequentially per timestep on the voxel grid, with predictions unprojected to mesh nodes via the precomputed nearest-voxel lookup. U-Net inference times in Table~\ref{tab:results} include voxelisation overhead.

\subsection{Data Organization for Machine Learning}

The final step converts the simulation outputs into machine-learning-ready samples without discarding the spatial structure that generated the thermal field. For each geometry variant, the dataset retains node coordinates, mesh connectivity, signed-distance values, process parameters, and time-resolved nodal temperatures. A temperature value without its spatial context is not enough for studying how geometry affects field prediction, and geometry without its corresponding transient field gives no learning target; by keeping both together, the dataset links what the casting looks like, how it is discretized, which process conditions it was simulated under, and how its temperature field evolves over time. Each case is additionally accompanied by a metadata record (geometry family, variant index, process parameters, split assignment, and file paths to the CAD, mesh, and field assets) so that cases can be queried and filtered without re-parsing raw solver output.

The data are partitioned at the geometry-variant level rather than at the individual timestep level, using 8 variants per family for training, 1 for validation, and 1 for test (an 80/10/10 split). Different timesteps from the same physical geometry cannot appear in both training and test sets, which prevents temporal samples from the same geometry from artificially inflating generalization performance.

\section{ML Surrogate Benchmarking}
\label{app:C}

To check whether learned surrogates can substitute for the FEM solve, two architectures, a point-wise multilayer perceptron (MLP) and a U-Net, were trained on the same dataset and evaluated against ground-truth simulation fields across all three test geometries (brake, flange, stepped block). For each geometry, three representative process-parameter variants were selected from the held-out set, labelled Best, Median, and Worst according to their surrogate-vs-ground-truth error ranking, and evaluated in two ways: (i) spatial absolute-error field maps at four representative time steps, rendered on a single shared colour scale per architecture-geometry pair for direct visual comparison of error severity, and (ii) the mean and maximum absolute error over the full simulated duration.

\subsection{Brake Geometry}

\begin{figure}[h]
  \centering
  \includegraphics[width=\linewidth]{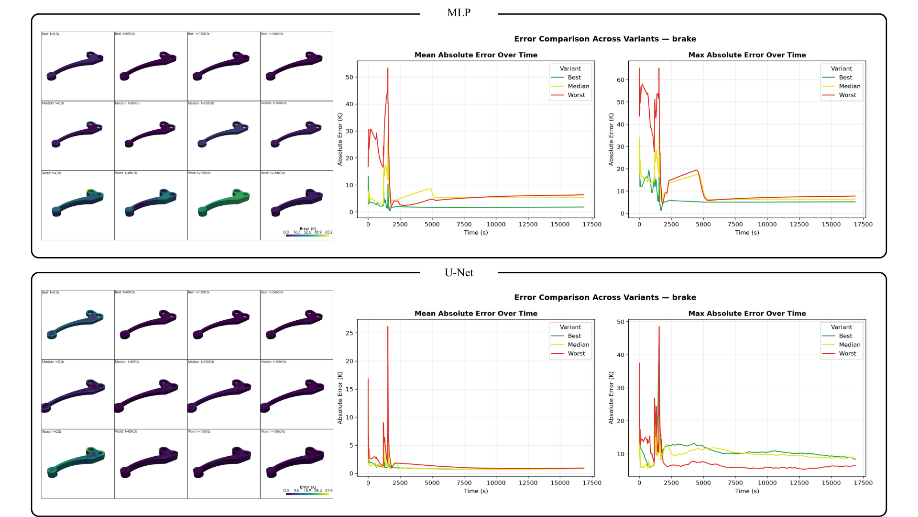}
  \caption{Temporal error evolution and spatial error distributions for MLP and U-Net predictions on the brake-lever geometry.}
  \label{fig:c1brake}
\end{figure}

For the Best, Median, and Worst test variants, the left panels show spatial absolute-error fields at representative time steps, while the middle and right panels report the evolution of mean and maximum absolute temperature error over the complete simulation interval. Both models exhibit their largest deviations during the early transient cooling period, when thermal gradients are strongest. The U-Net subsequently converges more rapidly toward low residual error, whereas the MLP retains larger errors for a longer portion of the transient response.

For the brake geometry (Figure~\ref{fig:c1brake}), both architectures exhibit their largest errors during the early transient period ($t < 2000$~s), coincident with the steepest thermal gradients following pouring. The MLP's Worst-case variant reaches a mean absolute error of approximately 53~K and a maximum of 65~K within this window, before decaying to a low, stable residual of 2-6~K by the end of the simulation. The U-Net shows a comparable transient error episode (peak mean error $\approx$ 26~K) but recovers markedly faster, converging to sub-2~K mean error well before the mid-point of the simulated interval. Field maps corroborate this: the MLP's error color scale spans 0-65.2~K across the shown snapshots, versus 0-37.9~K for the U-Net, with the U-Net's Worst-case panels visibly near-uniform (low-error) by the final time step. Both models preserve the expected Best/Median/Worst error ranking throughout, indicating that the models correctly capture relative case difficulty even where absolute accuracy differs.

\subsection{Flange Geometry}

\begin{figure}[h]
  \centering
  \includegraphics[width=\linewidth]{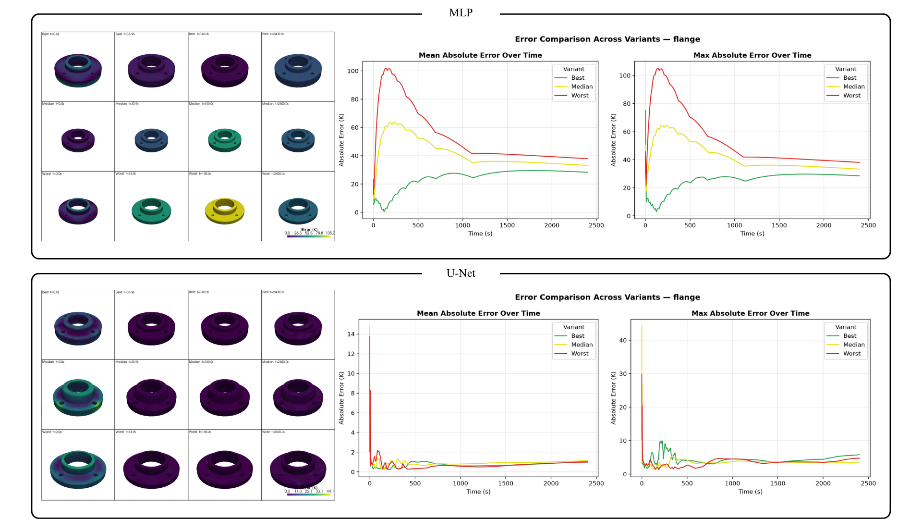}
  \caption{Temporal error evolution and spatial error distributions for MLP and U-Net predictions on the flange geometry.}
  \label{fig:c2flange}
\end{figure}

Spatial absolute-error fields are shown for Best, Median, and Worst test variants at representative times, together with the corresponding mean and maximum absolute temperature errors over the simulated cooling history. The MLP exhibits substantially larger and more persistent errors, particularly around the flange ring and bolt-hole regions where geometric discontinuities produce localized thermal gradients. In contrast, the U-Net error decreases rapidly after the initial transient and remains comparatively low throughout the later stages of cooling.

The flange geometry (Figure~\ref{fig:c2flange}) exposes a more substantial architectural gap. The MLP's mean absolute error for the Worst-case variant rises sharply to $\sim$100~K within the first 200~s and only partially relaxes thereafter, plateauing near 40~K for the remainder of the simulated window; that is, the model does not converge toward the low-error, near-isothermal condition expected at late-stage cooling. The Median-case variant shows the same qualitative behavior, plateauing near 35~K. By contrast, the U-Net exhibits only a brief, sharp error spike at $t \approx 0$ ($\approx$15~K, consistent with initialization/discretization transients) before collapsing to approximately 1~K mean error for all three variants by $t \approx 200$~s and remaining flat thereafter. The corresponding field maps show the MLP retaining visible high-error regions (yellow-green) around the bolt-hole and flange-ring discontinuities even at the final time step, while the U-Net's final-frame panels are uniformly low-error (dark) across all variants.

\subsection{Stepped Block Geometry}

\begin{figure}[h]
  \centering
  \includegraphics[width=\linewidth]{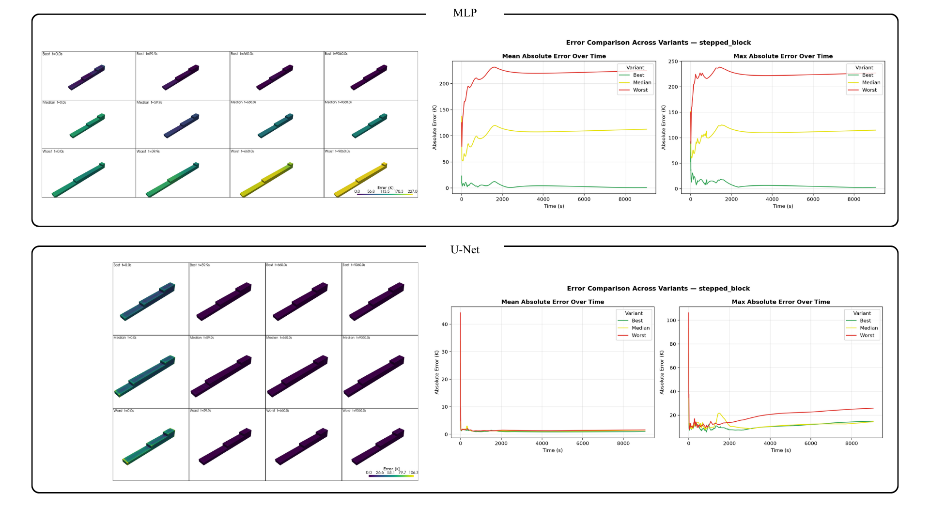}
  \caption{Temporal error evolution and spatial error distributions for MLP and U-Net predictions on the stepped-block geometry.}
  \label{fig:c3stepped}
\end{figure}

The figure compares spatial absolute-error fields and the temporal evolution of mean and maximum absolute errors for Best, Median, and Worst test variants. The stepped-block geometry produces the largest divergence between the two surrogate architectures, with the MLP developing a persistent error plateau associated with the abrupt cross-sectional transitions. The U-Net exhibits a substantially smaller transient error and converges to low mean error during the later cooling period, demonstrating the benefit of incorporating spatial context when predicting thermal fields across geometrically discontinuous regions.

The stepped-block results (Figure~\ref{fig:c3stepped}) show the largest absolute divergence between the two architectures. The MLP's Worst-case mean absolute error climbs to and plateaus near 230~K, with the Median-case similarly plateauing near 115~K, both well above any physically negligible residual, indicating the MLP fails to resolve the correct long-time thermal state around the geometry's cross-sectional step change. The U-Net, in contrast, again shows only an early, transient error spike ($\sim$45~K) before decaying to a sub-2~K mean error across all three variants by roughly $t = 500$~s; its maximum-error curve for the Worst-case variant settles near 20-25~K late in the simulation, an order of magnitude below the MLP's corresponding plateau.

\subsection{Discussion}

Across all three geometries, the U-Net surrogate consistently and substantially outperforms the MLP, both in peak transient error magnitude and, more importantly, in its ability to converge toward the correct steady/late-time thermal field rather than plateauing at a persistent bias. This effect is most pronounced for the flange and stepped-block geometries, both of which contain sharp geometric discontinuities (bolt holes and a ring structure in the flange; an abrupt cross-sectional step in the stepped block) that induce localized, spatially coupled thermal gradients. The brake geometry, which has a comparatively smooth, discontinuity-free organic shape, is the one case where the MLP's error, while still larger than the U-Net's, does decay to a low residual rather than plateauing, suggesting that the point-wise MLP's lack of spatial context becomes a critical limitation specifically when the underlying field has sharp local structure. The U-Net's convolutional receptive field, by comparison, appears able to propagate and reconcile local thermal gradients across neighbouring regions of the mesh, which is plausibly what allows it to track the correct late-time solidification state that the MLP misses. In all six architecture-geometry combinations, both models preserve the relative Best/Median/Worst difficulty ranking of the sampled variants, supporting the validity of the variant-selection procedure independent of which surrogate is used to evaluate it.

\end{document}